\documentclass[10pt,twocolumn,letterpaper]{article}

\usepackage{3dv}
\usepackage{times}
\usepackage{epsfig}
\usepackage{graphicx}
\usepackage{amsmath}
\usepackage{amssymb}
\usepackage{caption}
\usepackage{multirow}
\usepackage{wrapfig}
\usepackage{booktabs}
\usepackage{float}
\usepackage{textcomp}
\usepackage{xcolor}
\usepackage{bm}
\usepackage{makecell}
\usepackage[symbol]{footmisc}

\threedvfinalcopy %

\def\threedvPaperID{111} %

\ifthreedvfinal\fi

\usepackage{lipsum}

\newcommand\blfootnote[1]{%
  \begingroup
  \renewcommand\thefootnote{}\footnote{#1}%
  \addtocounter{footnote}{-1}%
  \endgroup
}

\begin{document}

\title{Controllable Radiance Fields for Dynamic Face Synthesis}

\author{Peiye Zhuang$^\text{1}$$^\star$$^\dagger$, Liqian Ma$^\text{2}$$^\dagger$, Sanmi Koyejo$^\text{1,3,4}$, Alexander Schwing$^\text{1}$\\
$^\text{1}$University of Illinois Urbana-Champaign, $^\text{2}$ZMO AI Inc., 
$^\text{3}$Stanford University,
$^\text{4}$Google Inc.\\
{\tt\small \{peiye, sanmi, aschwing\}@illinois.edu}, {\tt\small liqianma.scholar@outlook.com}
}

\twocolumn[{
\maketitle
    \begin{minipage}{\linewidth}
    \centering
    \vspace{-0.5cm}
    \includegraphics[width=\linewidth]{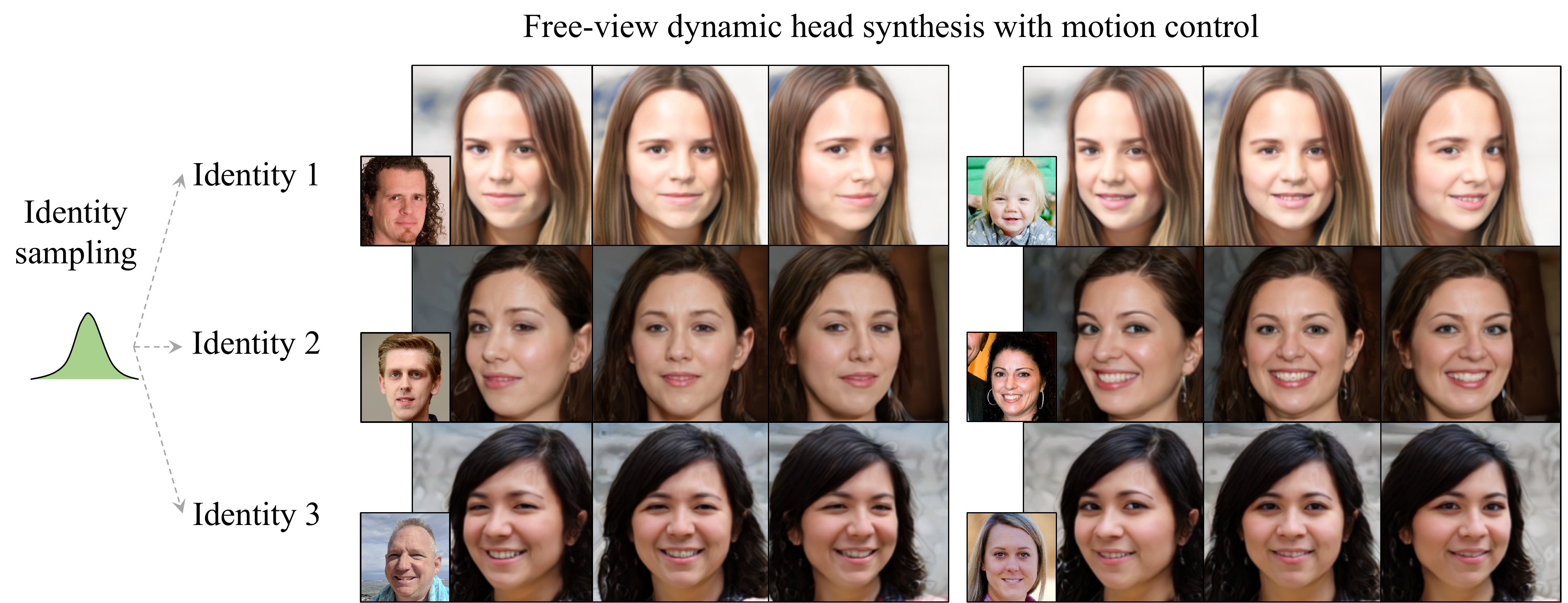}
    \end{minipage}
\vspace{-0.2cm}
    \captionof{figure}{\textbf{Controllable free-view dynamic head synthesis.} Each row presents an identity sampled from a prior distribution and two expressions guided by a reference image (\textit{bottom left}), viewed from multiple directions (\textit{column 1-3 and 4-6}).}
    \label{fig:1}
    \vspace{0.5cm}

}]
\thispagestyle{empty}

\begin{abstract}
\label{sec:01-abs}

\vspace{-.3cm}

Recent work on 3D-aware image synthesis has achieved compelling results using advances in neural rendering. 
However,  3D-aware synthesis of face dynamics hasn't received much attention. Here, we study how to explicitly control generative model synthesis of face dynamics exhibiting non-rigid motion (e.g., facial expression change), while simultaneously ensuring 3D-awareness. For this we propose a Controllable Radiance Field (CoRF): 1) Motion control is achieved by embedding motion features within the layered latent motion space of a style-based generator; 2) To ensure consistency of background, motion features and subject-specific attributes such as lighting, texture, shapes, albedo, and identity, a face parsing net, a head regressor and an identity encoder are incorporated. 
On head image/video data we show that CoRFs are 3D-aware while enabling editing of identity, viewing directions, and motion.

\blfootnote{$\star$ Some of the work was completed while P.Z. was at Google.}
\blfootnote{$\dagger$ Corresponding author.}

\end{abstract}
\vspace{-.2cm}
\section{Introduction}
\vspace{-.2cm}
\label{sec:01-intro}
 Face synthesis is an important task with applications in digital content creation, film-making and Virtual Reality (VR). Generative Adversarial Nets (GANs), as a powerful generative model, have demonstrated remarkable success in generating high-quality faces. However, despite the impressive performance of GANs in both 2D~\cite{goodfellow2014generative, karras2017progressive, brock2018large, karras2019style, karras2019analyzing, Karras2020ada} and 3D image synthesis~\cite{nguyen2019hologan, Schwarz2020NEURIPS, chanmonteiro2020pi-GAN,stylenerf}, its extension to \emph{both 3D-aware and motion-controllable} image synthesis has not been fully explored. 
This is largely due to the complexity of physical motion and the challenging yet consistent appearance changes of an identity. For example, dynamic head synthesis requires maintaining some 3D consistency over time to preserve facial identity while permitting other 3D deformations due to expression changes. It is even more challenging to enable interactive manipulation of identity, motion, and viewing directions.

Addressing this task of \textit{generalizable}, \textit{3D-aware}, and \textit{motion-controllable} synthesis of face dynamics enables  to generate never-before-seen faces including their dynamic motion,  while controlling the viewpoint, as shown in Fig.~\ref{fig:1}. %

Most related to this task are motion transfer methods such as face reenactment~\cite{Zakharov20, wiles2018x2face, siarohin2019animating, wang2019fewshotvid2vid, zakharov2019few, yang2018pose, siarohin2021motion, FOMM}. Specifically, prior work animates a subject shown in a source image given target motion from a driving video. For this, prior methods commonly use 2D GANs to produce dynamic subject images with additional guidance such as reference keypoints~\cite{Zakharov20, wiles2018x2face, siarohin2019animating, wang2019fewshotvid2vid, zakharov2019few, yang2018pose} and self-learned motion representations~\cite{FOMM, siarohin2021motion}.  However, due to intrinsic limitations of 2D convolutions, these approaches lack 3D-awareness. This leads to two main failure modes: 1) spurious identity changes -- the subject, head pose or facial expression of the source image is distinct from the target when rotating a source head by a large angle; 2) serious distortions when transferring motion across identity based on keypoints which contain identity-specific information such as face shapes. 
To address these failure modes, recent face reenactment works~\cite{ren2021pirenderer, gafni2020dynamic} estimate parameters of 3D morphable face models, e.g., poses and expression, as additional guidance.
Notably, Guy et al.~\cite{gafni2020dynamic} explicitly represent heads in a 3D radiance field showing impressive 3D-consistency. However, one model per face identity needs to be trained~\cite{gafni2020dynamic}. 

Different from this direction, we develop a generalizable method of motion-controllable synthesis for novel, never-before-seen identity generation. 
This differs from prior works which either consider motion control but drop 3D-awareness~\cite{TGAN2017, MoCoGAN, TGAN2020, Wang_2020_CVPR, Zakharov20, wiles2018x2face, FOMM, siarohin2019animating, wang2019fewshotvid2vid, zakharov2019few, yang2018pose}, or are 3D-aware but don't consider dynamics~\cite{Schwarz2020NEURIPS, chanmonteiro2020pi-GAN, stylenerf} or can't generate never-before-seen identities~\cite{ren2021pirenderer, gafni2020dynamic}.

To achieve our goal, we propose controllable radiance fields (CoRFs). We learn CoRFs from RGB images/videos with unknown camera poses. 
This requires to address two main challenges: 
1) how to effectively represent and control identity and motion in 3D; and 
2) how to ensure  spatio-temporal consistency across views and time.

To control identity and motion,  CoRFs use a style-based radiance field generator which takes  low-dimensional identity and motion representations as input, as shown in Fig.~\ref{fig:1}. 
Unlike prior head reconstruction work~\cite{ren2021pirenderer, gafni2020dynamic} that uses ground-truth images for self-supervision, there is no ground-truth target image for a generated image of a never-before-seen person. Thus, we propose an additional motion reconstruction loss to supervise motion control. 

To ensure spatio-temporal consistency, we propose three consistency constraints on face attributes, identities, and background. Specifically, at each training step, we generate two images given the same identity representation yet different motion representations.
We encourage that the two images share identical environment and subject-specific attributes. For this, we apply a head regressor and an identity encoder. A head regressor decomposes the images into representations of a statistic face model, including lighting, shape, texture and albedo. Then, we compute a consistency loss which compares the predicted attribute parameters of the paired synthetic images. Moreover, we use an identity encoder to ensure the paired synthetic images share the same identity. Further, to encourage a consistent background across time, we recognize background regions using a face parsing net and employ a background consistency loss between paired synthetic images that share the same identity representation.

As there is no direct baseline, we compare CoRFs to two types of work that are most related: 1) face reenactment work~\cite{FOMM, siarohin2021motion} that can control facial expression transfer; and 2) video synthesis work~\cite{MoCoGAN, TGAN2020, Wang_2020_CVPR, tian2021a} that can produce videos for novel identities. We evaluate the improvements of our method using the Fr\'echet Inception Distance (FID)~\cite{fid} for videos, motion control scores, and 3D-aware consistency metrics on FFHQ~\cite{karras2017progressive} and two face video benchmarks~\cite{facef, vox2} at a resolution of $256\times 256$. Moreover, we show additional applications such as novel view synthesis and motion interpolation in the latent motion space. 

\noindent\textbf{Contributions.} 1) We study the new task of \textit{generalizable}, \textit{3D-aware}, and \textit{motion-controllable} face generation.
2) We develop  CoRFs which enable editing of motion, and view-points of never-before-seen subjects during synthesis. For this, we study  techniques that aid motion control and spatio-temporal consistency.
3) CoRFs improve visual quality and spatio-temporal consistency of free-view motion synthesis compared to multiple baselines~\cite{MoCoGAN, TGAN2020, Wang_2020_CVPR, FOMM, Zakharov20, siarohin2021motion, tian2021a} on popular face benchmarks~\cite{karras2017progressive, facef, vox2} using multiple metrics for image quality, temporal consistency, identity preservation, and expression transfer.

\section{Related work}
\label{sec:02-related}

\begin{figure*}[t]
    \begin{minipage}{\linewidth}
    \centering
    \includegraphics[width=\linewidth]{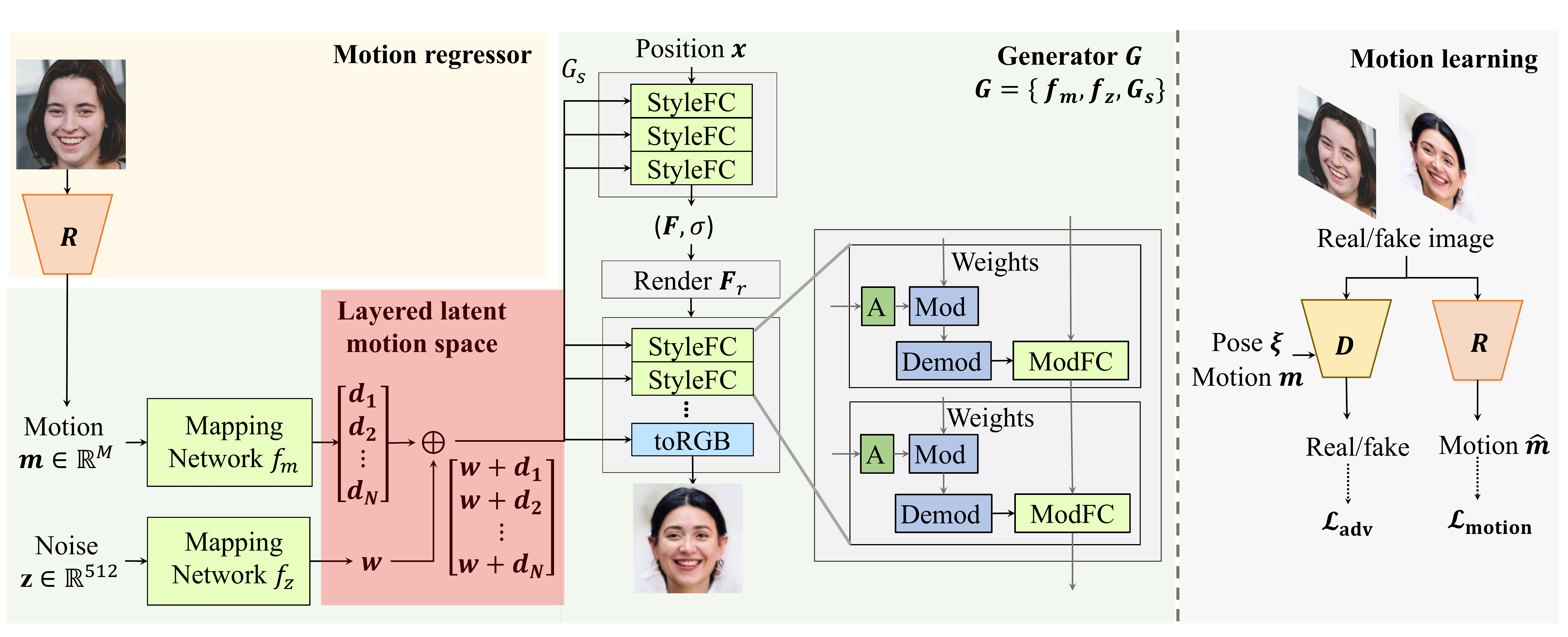}
    \end{minipage}
    \vspace{-0.3cm}
    \caption{\textbf{Overview of Controllable Radiance Fields (CoRFs).} In the offline step, we collect motion representations $\bm m$ using a regressor $R$. 
    At training time, the generator $G$ renders an image given position input $\bm x$, a noise $\bm z$, and a motion representation $\bm m$. The motion input $\bm m$ is embedded via a mapping network $f_m(\bm m)$ and combined with a style vector $\bm w$ via a summation ($\oplus$) before being used to generate the face. The discriminator $D$ compares real and generated images, conditioned on a camera pose $\bm \xi$ and a motion representation $\bm m$. To ensure visual quality and motion control, we employ a discriminator loss $\mathcal{L}_\text{adv}$ and a motion reconstruction loss $\mathcal{L}_\text{motion}$.
    }
    \vspace{-0.3cm}
    \label{fig:2}
\end{figure*}

\noindent\textbf{3D-aware image synthesis.} GANs~\cite{goodfellow2014generative} have significantly advanced 2D image synthesis capabilities in recent years, addressing early concerns regarding diversity, resolution, and photo-realism~\cite{karras2017progressive, brock2018large, karras2019style, karras2019analyzing, Karras2020ada, gulrajani2017improved}.  More recent GANs aim to extend 2D image synthesis to 3D-aware image generation~\cite{park2017transformation,  sun2018multiview, nguyen2018rendernet, nguyen2019hologan, Dosov3Dgeneration2017}, while permitting explicit camera control. 
For this,  methods~\cite{park2017transformation,  sun2018multiview, nguyen2018rendernet, nguyen2019hologan, Dosov3Dgeneration2017} use an implicit neural representation. 

Implicit neural representations~\cite{mildenhall2020nerf, chen2019learning, Mescheder_2019_CVPR, XuDISN2019, liu2019learning, sitzmann2019srns, sitzmann2019siren, park2019deepsdf, chabra2020deep} have been introduced to encode 3D objects (i.e., its shape and/or appearance) via a parameterized neural net. 
Compared to voxel-based~\cite{rezende2016unsupervised, Riegler_2017_CVPR, wu2016learning} and  mesh-based~\cite{pan2019deep, Wang_2018_ECCV} methods, implicit neural functions represent 3D objects in continuous space and are not restricted to a particular object topology. 
A recent variation, Neural Radiance Fields (NeRFs)~\cite{mildenhall2020nerf}, represents the appearance and geometry of a \textit{static} real-world scene with a multi-layer perceptron (MLP), enabling impressive novel-view synthesis with multi-view consistency via volume rendering. Different from the studied method, those classical approaches can't generate  novel scenes.

To address this, recent work~\cite{Schwarz2020NEURIPS, chanmonteiro2020pi-GAN, stylenerf, zhou2021CIPS3D,ZhaoECCV2022a} studies 3D-aware generative models using NeRFs as a generator.
For instance, GRAF~\cite{Schwarz2020NEURIPS} and $\pi$-GAN~\cite{chanmonteiro2020pi-GAN} operate on a randomly sampled camera pose and latent vectors from prior distributions to produce a radiance field via an MLP. StyleNeRF and CIPS-3D~\cite{stylenerf,zhou2021CIPS3D} combine a shallow NeRF network to provide low-resolution radiance fields and a 2D rendering network to produce high-resolution images with fine details. LolNeRF~\cite{rebain2022lolnerf} learns 3D objects by optimizing foreground and background NeRFs together with a learnable per-image table of latent codes. Zhao \etal~\cite{ZhaoECCV2022a} develop a generative multi-plane image (GMPI) representation to ensure view-consistency. 
These works only generate multi-view images for \textit{static} scenes. In contrast, the proposed CoRF targets generation of \textit{face dynamics} while enabling to control motion.
We also notice concurrent NeRF-based generative models~\cite{sun2022fenerf, hong2021headnerf} aiming for 3D-aware semantic appearance or expression editing.

\noindent\textbf{Dynamic object synthesis.} Dynamic object synthesis has been studied using 2D- and 3D-based methods. Some 2D-based methods unconditionally generate dynamic instances via convolutional neural nets such as GANs~\cite{vgan, TGAN2017, TGAN2020, MoCoGAN, Wang_2020_CVPR}, but struggle with motion control. 
Follow-up work incorporates additional reference information during synthesis~\cite{Zakharov20, wiles2018x2face, FOMM, siarohin2019animating, wang2019fewshotvid2vid, zakharov2019few, wang2021facevid2vid, ren2021pirenderer}, e.g., for tasks like \textit{face reenactment}. Our method is related but differs in that CoRF synthesizes images and controls motion for \textit{never-before-seen} identities from \textit{free viewpoints}. 
3D-based methods recover face geometry and control the expression by changing geometry parameters~\cite{thies2016face2face, kim2018deep, geng2018warp, nagano2018pagan}. While they can control facial motion effectively, they struggle to produce realistic hair, teeth, and accessories~\cite{geng2018warp, nagano2018pagan}. 
Recent work~\cite{gafni2020dynamic} adapts neural implicit methods, e.g., NeRFs to learn head reconstruction from images. However,  one model is trained per identity, i.e., these methods don't generalize across different identities. Unlike prior work~\cite{gafni2020dynamic}, CoRF generalizes across identities.

\noindent\textbf{Motion conditioning.} We draw inspiration from GAN-based image inversion and editing work~\cite{abdal2019image2stylegan, zhuang2021enjoy}. To be concrete, Abdal et al.~\cite{abdal2019image2stylegan} proposed to embed images back to an extended latent identity space of a StyleGAN~\cite{karras2019style}, i.e.,  the $W^+$ space. This extended latent identity space was shown to be remarkably useful for image editing and expression transfer in  follow-up work~\cite{zhuang2021enjoy}. In our case, we embed a motion representation into a layered latent motion space, which we then broadcast to every synthesis layer in the generator. In Section~\ref{sec:04-exp}, we illustrate that such a motion embedding strategy benefits motion control. 

\noindent\textbf{3D consistency preservation.} To ensure 3D consistency over time, a 3D convolutional discriminator is commonly used to distinguish the source of input video clips~\cite{tian2021a}. Recent work~\cite{yu2022generating} finds that when employing an implicit neural rendering net as a generator, a 2D discriminator for 2 synthetic frames in a video is sufficient. Inspired by this result, we propose consistency losses on paired synthetic frames without using a 3D convolutional discriminator. 

\section{Method}
\label{sec:03-method}

We aim for generalizable, 3D-aware and motion-controllable synthesis of face dynamics. Concretely, we seek to generate never-before-seen faces while controlling dynamic motions and viewpoint arbitrarily.
For this, we propose controllable radiance fields (CoRF), illustrated in Fig.~\ref{fig:2}, for which we provide an overview next.

\noindent\textbf{Overview.} CoRFs are composed of a generator $G$, a discriminator $D$, a regressor $R$ and an identity encoder $E$.
The CoRF generator $G$ renders an RGB image $I'$ based on %
a noise vector $\bm z$ and a motion representation $\bm m$.
Concretely, the generator $G$ first applies two mapping modules $f_z$ and $f_m$, before rendering the image via a style-based synthesis module $G_s$, i.e., $G = \{f_z, f_m, G_s \}$. Specifically, to ensure effective motion control, we embed motion representations $\bm m$ into a layered latent motion space using a mapping net $f_m$, which is then combined with a style vector $\bm w:= f_z(\bm z)$ via a summation. 
The discriminator $D$ is applied to ensure photo-realism. 
The regressor $R$ and the encoder $E$ ensure spatio-temporal consistency of synthetic faces over dynamic changes.

During training, different from prior work, we generate two images given the same identity yet different motion representations. 
To preserve consistency over views and poses, the regressor $R$ and the identity encoder $E$ extract features for the paired images which are encouraged to be similar via a consistency loss $\mathcal{L}_\text{consist}$ and an identity loss $\mathcal{L}_\text{id}$. 
To supervise motion control, we estimate the motion representation from images and incorporate a motion reconstruction loss $\mathcal{L}_\text{motion}$, to minimize the distance between the estimated motion and the conditioned one. Finally, we apply a background loss, $\mathcal{L}_\text{bg}$, to preserve background consistency over dynamic changes.

In the following, we briefly describe the generator $G$ and the discriminator $D$ (Section~\ref{sec:3-1}). We then describe our contributions: 1) how to control motion (Section~\ref{sec:3-3}) and 2) how to preserve spatio-temporal consistency (Section~\ref{sec:3-4}).

\subsection{Generator and Discriminator}
\label{sec:3-1}

\noindent\textbf{Generative radiance fields.} 
We adopt the generative strategy of StyleNeRF~\cite{stylenerf} where a style-based radiance field generator $G$ renders a synthetic image $I'$. Specifically, we render colors for each pixel from an integral feature $\bm F_r$. To obtain an integral feature $\bm F_r$, we integrate features $\bm F$ of 3D points along a camera ray $\bm r$. We present the generator $G$ and its workflow in Fig.~\ref{fig:2}.

Formally, we cast rays from the camera origin $\bm o$ through the pixels of the image plane, and accumulate the sampled density $\sigma$ and feature values $\bm F$ along the rays with near and far planes $t_n$ and $t_f$. 
The aggregated feature $\bm F_r$ is computed using the rendering equation~\cite{max1995optical} as follows:
\begin{eqnarray}
    &&\bm F_r = \int^{t_f}_{t_n} T(t) \sigma(\bm r(t) \bm F (\bm r(t), \bm d, \bm m, \bm z)) dt, \nonumber \\
    &&\text{where } T(t) = \exp\left(-\int^{t}_{t_n} \sigma(\bm r(s), \bm m, \bm z)ds \right).
\end{eqnarray}
Here, $\bm d$ refers to the ray direction and $T(t)$ corresponds to the accumulated transmittance along the ray $\bm r$ from $t_n$ to $t$. The functions $\sigma(\bm r(t), \bm m, \bm z)$ and $\bm F (\bm r(t), \bm d, \bm m, \bm z)$ are the density and the feature at a 3D location, respectively. 

Different from StyleNeRF~\cite{stylenerf}, the generator $G$ in CoRF is conditioned on a motion representation $\bm m \in \mathbb{R}^{50}$ for motion control. To this end, a ReLU mapping network $f_m$ embeds the motion representation $\bm m$ into a latent space, which is then injected to the synthesis module $G_s$ via weight modulation~\cite{karras2019style}. We discuss the details in Section~\ref{sec:3-3}.
Also  different from StyleNeRF~\cite{stylenerf}, the generator $G$ does not contain upsampling blocks. Instead, we increase the image resolution by sampling more rays.

\noindent\textbf{Discriminator.}
We use the estimated camera pose $\bm \xi$ and the motion feature $\bm m$ of a training image $I\sim p_\text{real}$ to control generation of new images $I' \sim p_\text{syn}$. The discriminator $D$ is conditioned on a camera pose $\bm \xi$ and a motion feature $\bm m$, following the conditional strategy of StyleGAN2-ADA~\cite{Karras2020ada, Chan2021}. 
We use a non-saturating adversarial loss with a gradient regularizer~\cite{mescheder2018training} applied  on the discriminator $D$. We refer to this loss as $\mathcal{L}_\text{adv}$, i.e., 
\begin{align}
    \mathcal{L}_\text{adv} &= \mathbb{E}_{I' \sim p_\text{syn}, \bm m \sim p_m, \bm \xi \sim p_{\xi}} [ f(D (I', \bm \xi, \bm m)) ] \nonumber \\
    & + \mathbb{E}_{I \sim p_\text{real}} [ f(-D(I)) + \Vert \nabla D(I) \Vert^2], \nonumber
\end{align}
where $f(u) \triangleq - \log (1 + \exp(-u))$. Further, $ p_m$ and $p_\xi$ refer to  distributions over motion vectors and pose respectively. In practice, we extract  motion features and pose  from training data samples using~\cite{DECA}. During training, we randomly sample a training image with its corresponding estimated motion feature and pose to control image synthesis.

\subsection{Motion control}
\label{sec:3-3}
A key goal of CoRF is to control synthesis of 3D-aware face dynamics. 
For this, we condition the generator $G$ on low-dimensional motion representations $\bm m$. In this section, we discuss our conditioning method, motion extraction from images, and the supervision which we use during training to encourage motion control.

\noindent\textbf{Motion conditioning.} We embed the motion representation in a layered latent motion space using the mapping module $f_m$. The layered latent motion space is a concatenation of $N$ latent motion vectors, denoted as $[\bm{d_1}, \dots,\bm{d_N}]^T := f_m(\bm m)$, where $\bm{d_i} \in \mathbb{R}^{512}$ for  $i \in \{1,\dots,N \}$. %
Here, $N$ corresponds to the number of layers in the synthesis module $G_s$, which we find to lead to feature embeddings with superior performance. %
The latent motion vectors are then combined with the latent style vector $\bm w$. Formally, we combine style vector and motion representation via a summation, i.e., $$[\bm w + \bm{d_1}, \dots,\bm w + \bm{d_N}]^T.$$
The conditioned latent vectors are then given to each layer of the synthesis module $G_s$ via modulation~\cite{karras2019style}.

Instead of summation, other conditioning strategies for control exist. However, we find that not every strategy works equally well in our task. For example, recent head reconstruction work~\cite{gafni2020dynamic} concatenates a motion representation with a noise vector $\bm z$ which is then provided as input to the first layer of the model. We compare our conditioning strategy with this prior method~\cite{gafni2020dynamic} in Section~\ref{sec:04-exp} and find that our motion conditioning strategy  accelerates  convergence of the motion reconstruction loss $\mathcal{L}_\text{motion}$.

\noindent\textbf{Efficient motion extraction.} To obtain a low-dimensional motion representation $\bm m$ for face dynamics, we extract coefficients of a blendshape basis of a statistical head model using a pre-trained regressor $R$ with an accurate statistical head model~\cite{FLAME}. For this we use a publicly available package.\footnote{Code available from https://github.com/YadiraF/DECA}

Note that our method is also compatible with other pre-trained head models. Compared with existing work~\cite{gafni2020dynamic} that uses an \textit{optimization-based} approach for expression estimation, use of a pre-trained regressor $R$ enjoys two advantages. First, using the regressor $R$ is computationally more efficient as coefficient prediction requires only a single forward-pass. Second, we can use the regressor $R$ as a differentiable plug-in to regress and supervise the expression representation of synthetically generated frames. 

\noindent\textbf{Motion learning.}
A motion reconstruction loss $\mathcal{L}_\text{motion}$ is applied to supervise motion control. It assesses whether the synthetically generated image exhibits the motion for which the representation was provided as input. For this we use a simple quadratic loss, i.e., 
\begin{align}
\mathcal{L}_\text{motion}  = \mathbb{E}_{I' \sim p_\text{syn}} \Vert \bm m - \hat{\bm m} \Vert^2,
\end{align}
where $\bm{\hat{m}}$ is estimated by the regressor $R$, i.e., $\bm{\hat{m}} = R(I')$.

\begin{figure}[t]
    \centering
    \includegraphics[width=\linewidth]{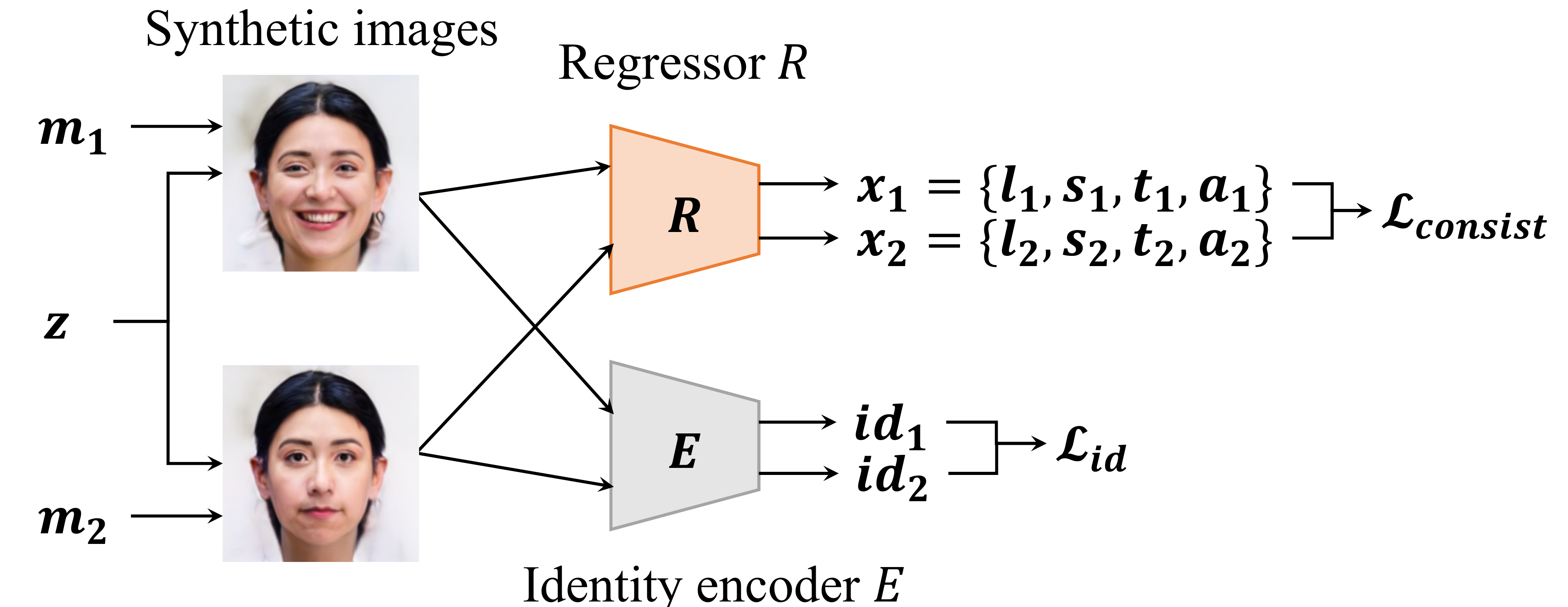}
    \caption{\textbf{Consistency learning.} We propose two losses, $\mathcal{L}_\text{consist}$ and $\mathcal{L}_\text{id}$, to ensure spatio-temporal consistency of paired generated images over dynamics, which includes lighting ($\bm l$), texture ($\bm t$), shape ($\bm s$), albedo ($\bm a$), and identity ($\bm{id}$).}
    \label{fig:3}
\end{figure}

\subsection{Spatio-temporal consistency}
\label{sec:3-4}

\begin{figure*}[t]

    \begin{minipage}{\linewidth}
    \includegraphics[width=\linewidth]{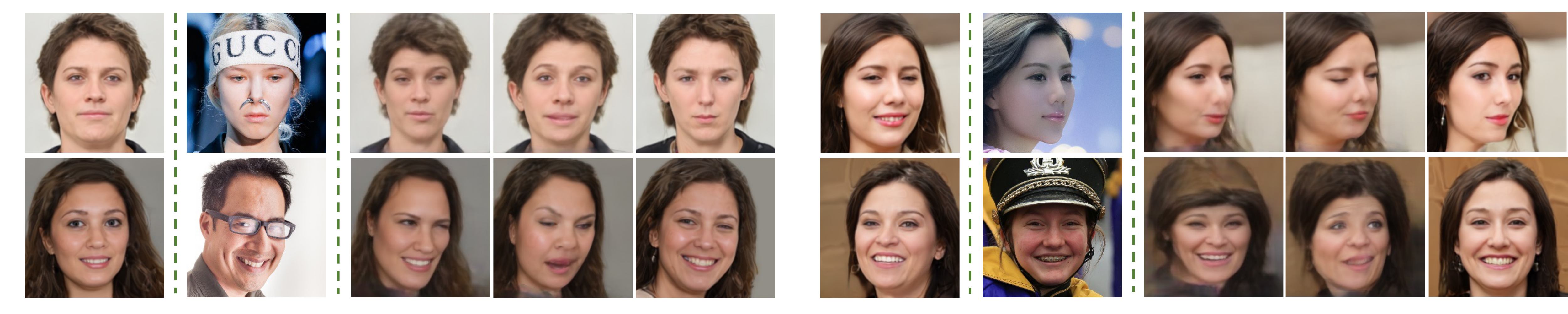}
    \end{minipage}
    
    \begin{minipage}{\linewidth}
   \hspace{.4 cm} Source \hspace{.5 cm}  Driving \hspace{.3cm}  FOMM~\cite{FOMM} \hspace{.1cm}  MRAA~\cite{siarohin2021motion} \hspace{.1cm}  Ours
   \hspace{.8 cm} 
   Source \hspace{.5 cm}  Driving \hspace{.2 cm} FOMM~\cite{FOMM} \hspace{.1cm}  MRAA~\cite{siarohin2021motion} \hspace{.1cm}  Ours
   
    \end{minipage}
    \vspace{-0.5cm}
    \caption{\textbf{Visual comparison of face reenactment on FFHQ.} We transfer expression and  pose from  driving images to the source images. We compare our results with results of FOMM~\cite{FOMM} and MRAA~\cite{siarohin2021motion} at a resolution of  $256\times 256$ in challenging  cases. We show face reenactment across identity, gender, pose, or with a partially occluded driving image. }
    \label{fig:6}
    \vspace{-0.3cm}
\end{figure*}

The aforementioned losses are insufficient to ensure spatio-temporal consistency. To address this, we propose two losses which focus on the generated face and one loss which focuses on background regions. All three losses are simple yet effective at preserving spatio-temporal consistency. 
In summary, to optimize CoRF we minimize the weighted objective 
\begin{equation}
      \mathcal{L} \triangleq  \lambda_1 \mathcal{L}_\text{adv} + \lambda_2 \mathcal{L}_\text{motion} + \lambda_3 \mathcal{L}_\text{consist} +\lambda_4 \mathcal{L}_\text{id} + \lambda_5 \mathcal{L}_\text{bg},
\end{equation}
where the hyper-parameters $\lambda_1, \lambda_2, \ldots, \lambda_5$ adjust the strength of the different loss terms. In practice, the regressor $R$ and the identity encoder $E$ are pre-trained offline on a large-scale dataset~\cite{DECA, cao2018vggface2}. We describe each of the three additional losses $\mathcal{L}_\text{consist}$, $\mathcal{L}_\text{id}$ and $\mathcal{L}_\text{bg}$ next.

\noindent\textbf{Consistency of attributes.} 
The motion conditioning occasionally changes lighting and person-specific attributes such as albedo and shape, resulting in flickering textures in rendered images. To alleviate this issue, we propose a consistency loss to supervise  attributes independent from motion (see Fig.~\ref{fig:3}). 
Specifically, we extract lighting, texture, shape, and albedo representations for paired artificially generated images which have an identical noise vector $\bm z$, yet a different motion representation $\bm m$. These extracted attributes are independent from motion and thus should be identical for paired  images. To encourage their consistency via a loss, we subsume representations of these attributes (lighting, albedo, texture, shape) for both paired images in the set ${\cal A} = \{(\bm l_1, \bm l_2), (\bm a_1, \bm a_2), (\bm t_1, \bm t_2), (\bm s_1, \bm s_2) \}$. We then compute the consistency loss
\begin{align}
    \mathcal{L}_\text{consist} = \mathbb{E}_{I' \sim p_\text{syn}} \left[\sum_{(\bm x_1, \bm x_2) \in \mathcal{A}} \beta_x \Vert \bm x_1 - \bm x_2 \Vert^2\right],
\end{align}
by comparing pairs of estimated attributes $\bm x_1$ and $\bm x_2$. Here, $\bm l, \bm a, \bm t$, and $\bm s$ are the attributes of lighting, albedo, texture, and shape, respectively. $\beta_x$ is the loss weight. In practice, these attributes are predicted by the regressor $R$. %

\noindent\textbf{Consistency of identity.} Beyond attribute consistency, we use an identity loss on paired images, following  prior face reconstruction work~\cite{DECA} (see Fig.~\ref{fig:3}). To be concrete, an encoder $E$ outputs identity embeddings of  paired generated images which have an identical noise vector $\bm z$ yet a different motion representation. Let $I_1'$ and $I_2'$ refer to the two generated images. The identity loss is proportional to the negative cosine similarity between embeddings of both images, i.e., it is computed via 
\begin{align}
    \mathcal{L}_\text{id}  = \mathbb{E}_{I_1' \sim p_\text{syn}, I_2' \sim p_\text{syn}} \left[1 - \frac{E(I_1') \cdot E(I_2')}{ \Vert E(I_1') \Vert_2 \cdot \Vert E(I_2')\Vert_2}\right].
\end{align}

\noindent\textbf{Consistency of background.} Beyond the constraints on faces, we also want to encourage that the background remains mostly static over time. For this, we recognize background regions of paired synthetic images $I_1'$ and $I_2'$, using a pre-trained face parsing net.\footnote{Code available from https://github.com/zllrunning/face-parsing.PyTorch} The corresponding binary background masks, which we refer to via $s_1$ and $s_2$, use a value of $1$ to indicate the background regions and a value of $0$ to indicate the face regions. Using both masks, we define a background consistency loss on the overlapping background regions of paired images via
\begin{align}
    \mathcal{L}_\text{bg}  = \mathbb{E}_{I_1' \sim p_\text{syn}, I_2' \sim p_\text{syn}} 
     \left[\frac{ \Vert (s_1 \cdot s_2) \cdot (I_1' - I_2') \Vert_2}{\Vert s_1 \cdot s_2 \Vert}\right],
\end{align}
where $\cdot$ denotes an element-wise product. Due to the limited space, we illustrate details of the background loss $\mathcal{L}_\text{bg}$ in the supplementary material.

\section{Experiments}
\label{sec:04-exp}

\subsection{Experimental settings}
\noindent\textbf{Datasets and preprocessing.} We evaluate the proposed approach on FFHQ~\cite{karras2017progressive} and two face video datasets, including Faceforensics++~\cite{facef} and VoxCeleb2~\cite{vox2}. (i) \textbf{FFHQ}~\cite{karras2017progressive}: We show CoRF learning motion dynamics on FFHQ, which is a single image dataset. (ii) \textbf{The Faceforensics++ dataset}~\cite{facef} contains 1,000 raw video sequences. 
(iii) \textbf{The VoxCeleb2 dataset}~\cite{vox2} contains over 1 million videos for 6,112 identities. 
As videos in these two face datasets~\cite{facef, vox2} contain low-quality frames with small or occluded faces, a pre-processing step is applied to clean the data. Specifically, we detect the faces in the videos using a pretrained landmark detection model~\cite{bulat2017far}, sharpen faces, filter out frames with small, blurry, or occluded faces, and align the remaining faces. We finally use 990 and 4,804 videos in each dataset~\cite{facef, vox2} for CoRF training.

\begin{figure}[t]
    \vspace{-.3cm}
     \begin{minipage}{\linewidth}
    \includegraphics[width=\linewidth]{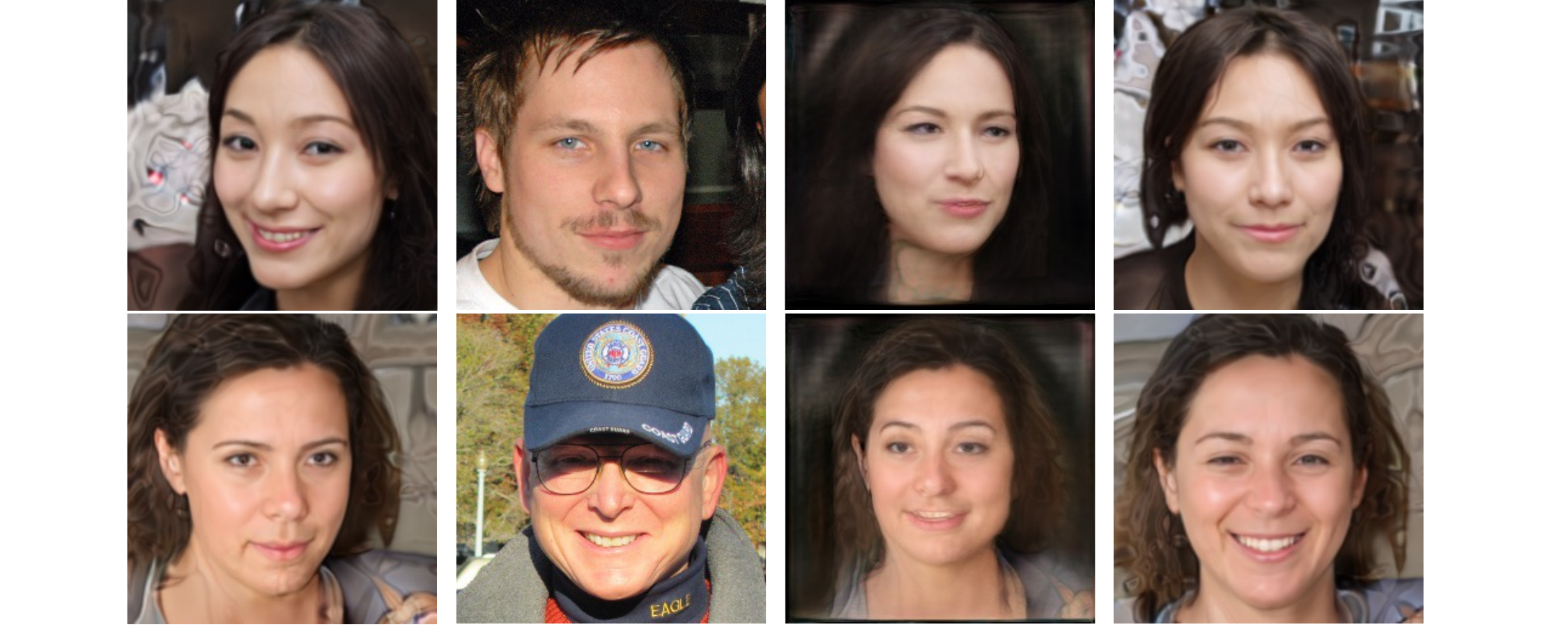}
    \end{minipage}
    
    \vspace{.1cm}
    \begin{minipage}{\linewidth}
    \hspace{1.cm} Source 
    \hspace{.3cm} Driving
    \hspace{.3cm} Bi-layer~\cite{Zakharov20} 
    \hspace{.3cm}  Ours  
    \end{minipage}

    \vspace{-.2cm}
    \caption{\textbf{Visual comparison to Bi-layer~\cite{Zakharov20}.}}
    \label{fig:b1}
    \vspace{-0.1cm}
\end{figure}

\noindent\textbf{Baselines and evaluation metrics.} 
We evaluate CoRF from two perspectives.
First, we quantitatively compare our CoRF with face reenactment work~\cite{FOMM, tian2021a}. To this end, we use the Fréchet Inception Distance (FID)~\cite{fid} and 3 cosine similarity scores for identity preservation and expression transfer, namely ID, EXP$^\dag$ and EXP$^\ddag$.
Regarding identity preservation, we extract face features using a popular image identity recognition mode pretrained on VGGface2~\cite{cao2018vggface2}. 
Regarding expression transfer, for a fair evaluation, we use another pretrained facial expression estimation network~\cite{deng2019accurate} (EXP$^\ddag$), which differs from the encoder~\cite{DECA} (EXP$^\dag$) used in our model.
Second, as CoRF generates never-before-seen dynamic faces from a prior noise distribution, similar to unconditional video synthesis approaches~\cite{MoCoGAN, TGAN2020, Wang_2020_CVPR, tian2021a}, we compare CoRF with these baselines using the Fréchet Video Distance (FVD)~\cite{fvd} score. The FVD score is a video-level FID score commonly used to examine both visual quality and temporal coherence of synthesized dynamics~\cite{Wang_2020_CVPR, hara2018can}.

\subsection{Results}

\begin{figure}[t]

    \begin{minipage}{0.13\linewidth}
    \centering
    \vspace{.2cm}
    Driving
    
    \vspace{.7cm}
    Ours
    
    \vspace{.6cm}
    Driving
    
    \vspace{.7cm}
    Ours
    \end{minipage}
    \centering
     \begin{minipage}{0.85\linewidth}
    \includegraphics[width=\linewidth]{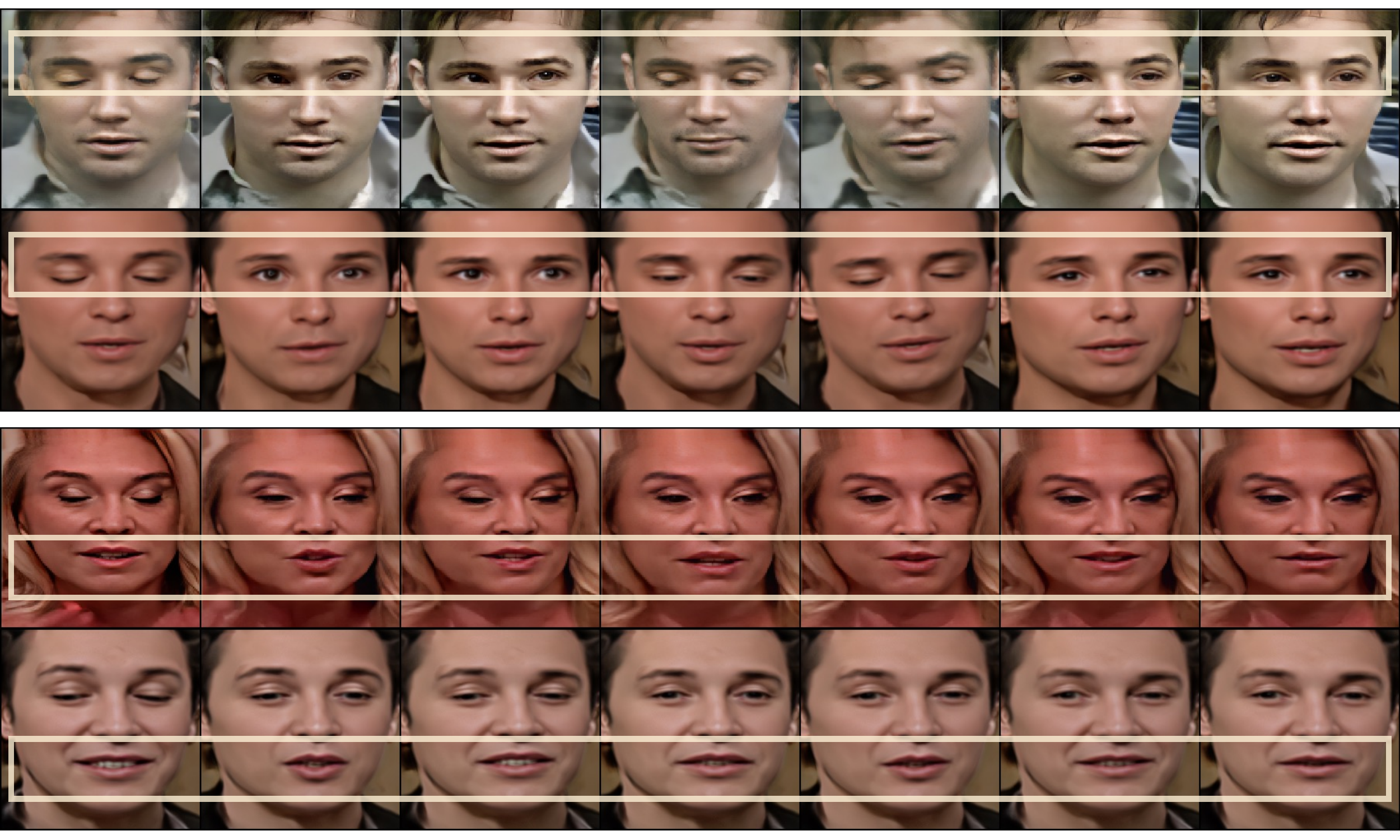}
    \end{minipage}
    \caption{\textbf{Targeted video generation on VoxCeleb2 data.} We generate head videos (\textit{row 2 and 4}) conditioned on facial expression information of the driving frames (\textit{row 1 and 3}) at a $256 \times 256$ resolution. Note, the synthetic portraits have a similar expression as the driving frame (see highlighted eyes in row 2, and the mouth in row 4).}
    \label{fig:7}
\end{figure}

\noindent\textbf{Comparison to face reenactment methods.}
We present a visual comparison to FOMM~\cite{FOMM}, MRAA~\cite{siarohin2021motion}, and Bi-layer~\cite{Zakharov20} in Fig.~\ref{fig:6} and Fig.~\ref{fig:b1}. For this, we transfer expressions and poses from real `driving' images to synthetic source images in four challenging cases, including cross-identity, cross-gender, cross-pose, and driving image occlusion. %
We observe that the baseline methods struggle in these cases, while  CoRF better preserves face geometry and identity. Additionally, we show  targeted video generation on the VoxCeleb2 dataset~\cite{vox2} at a $256\times 256$ resolution in Fig.~\ref{fig:7}. 
Different from the face reenactment baselines~\cite{FOMM, siarohin2021motion}, CoRF is capable of controlling more factors such as novel views and field-of-view (FOV) without the requirement of a reference image, e.g., in Fig.~\ref{fig:fov_main}. We show additional results in the supplementary material.

In Tab.~\ref{tab:1}, we quantitatively compare face reenactment methods on 10,000 pairs of source and driving images (variances in the supplementary material).  We find CoRF to significantly improve upon the baselines~\cite{FOMM, siarohin2021motion} with regards to identity preservation and expression transformation. 

\begin{figure}[t]
    \centering
     \begin{minipage}{\linewidth}
    \includegraphics[width=\linewidth]{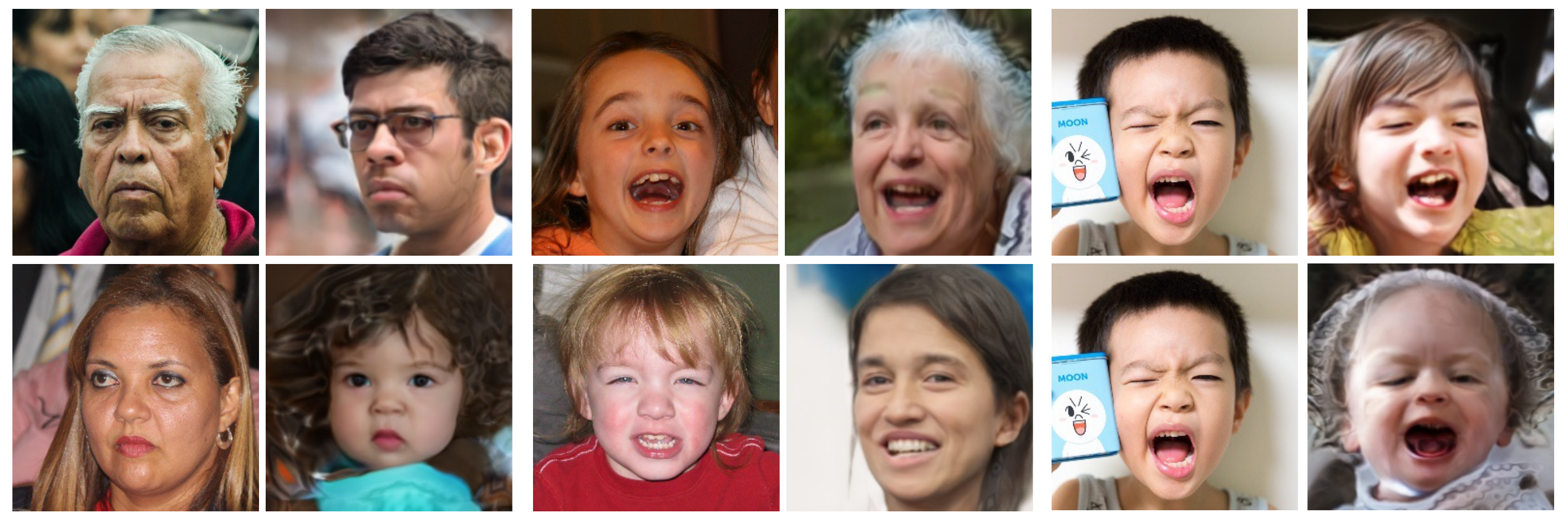}
    \end{minipage}
    
    \vspace{.1cm}
    \begin{minipage}{\linewidth}
    \hspace{0.2cm} Driving \hspace{.2cm} Result \hspace{.2cm} Driving \hspace{.2cm} Result \hspace{.2cm} Driving \hspace{.2cm} Result
    \end{minipage}

    \caption{\textbf{Rendered results with complex driving expressions.}}
    \label{fig:expression}
    \vspace{-0.3cm}
\end{figure}

\noindent\textbf{Complex expression animation.} 
 We show complex expression animation in Fig.~\ref{fig:expression}. Among all randomly generated images, we observe the results of the proposed method to follow the expressions in the driving image, highlighting the method's capability.

\begin{figure}[h]
    \centering

    \begin{minipage}{\linewidth}
     \centering
    \includegraphics[width=\linewidth]{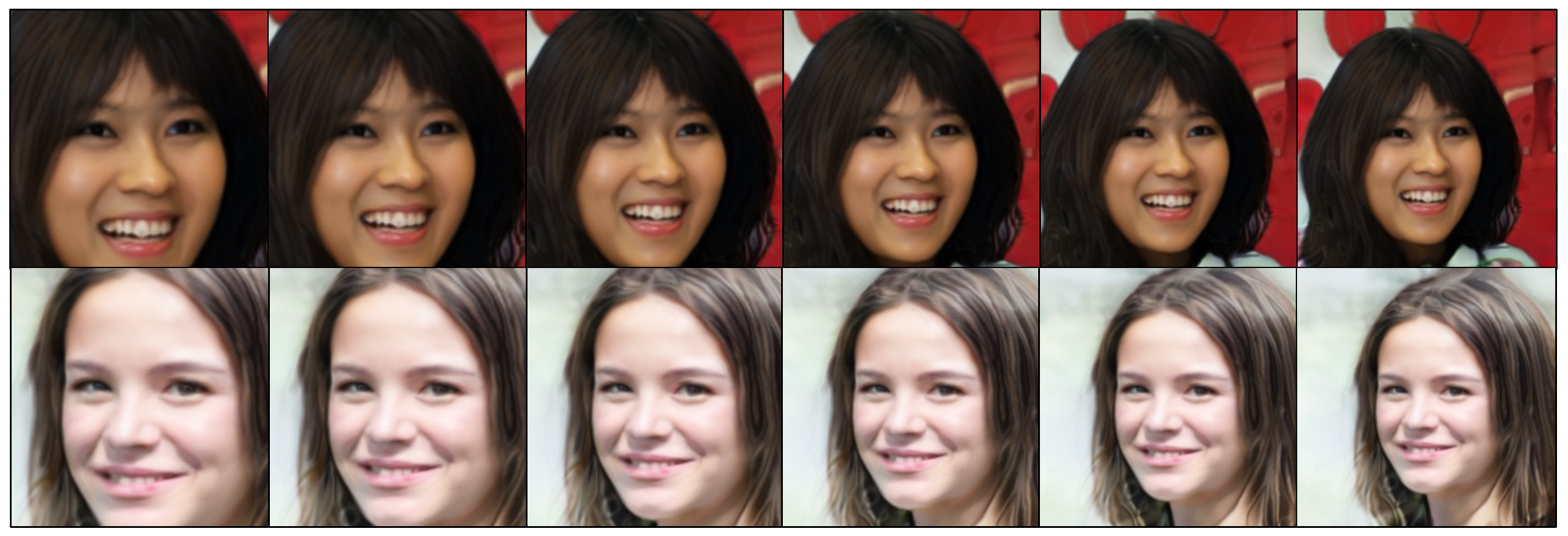}
    \end{minipage}
    
    \caption{\textbf{Rendered results under camera zoom-in to zoom-out.}}
    \label{fig:fov_main}
\end{figure}

\begin{table}[t]
    \centering
    \setlength{\tabcolsep}{.6pt}{
    \begin{tabular}{cccccccc}
    \toprule
    \centering
    ~ &~ &FaceF~\cite{facef} &~&~ &~ &Vox2~\cite{vox2} &~ \\
    \cmidrule(r){2-5} 
    \cmidrule(r){6-8} 
    ~~~~~~~~~~\qquad\qquad &~\cite{FOMM} &~\cite{siarohin2019animating} &Ours &~~~~&~\cite{FOMM} &~\cite{siarohin2019animating} &Ours \\
    \midrule
    FID $\downarrow$ & 24.44   &20.62 &\textbf{1.53}  \qquad &~\qquad
                     &32.01   &26.26 &\textbf{3.64}\\
    ID $\uparrow$    & 0.934    &0.965 &\textbf{0.973} \qquad &~\qquad
                    &0.942 &0.968  &\textbf{0.988} \\
    EXP$^\dag$ $\uparrow$ &0.592 &0.763 &\textbf{0.815} \qquad &~\qquad
                    & 0.748 &0.837  &\textbf{0.938}\\
    EXP$^\ddag$ $\uparrow$ &0.543 &0.595 &\textbf{0.601} \qquad &~\qquad
                    & 0.459  &0.481 &\textbf{0.731} \\
   
     \bottomrule
    \end{tabular} 
    }
    \caption{\textbf{Quantitative evaluation for face reenactment.} We compare our method with FOMM~\cite{FOMM} and MRAA~\cite{siarohin2019animating} at a $256 \times 256$ resolution. We compute the FID score~\cite{fid}, the face identity preservation score (ID), and expression similarities (two expression extractors~\cite{DECA} and~\cite{deng2019accurate} are used, denoted as EXP$^\dag$ and EXP$^\ddag$, respectively).}
    \vspace{-0.3cm}
    \label{tab:1}
\end{table}

\begin{figure}[t]
\centering
\includegraphics[width=\linewidth]{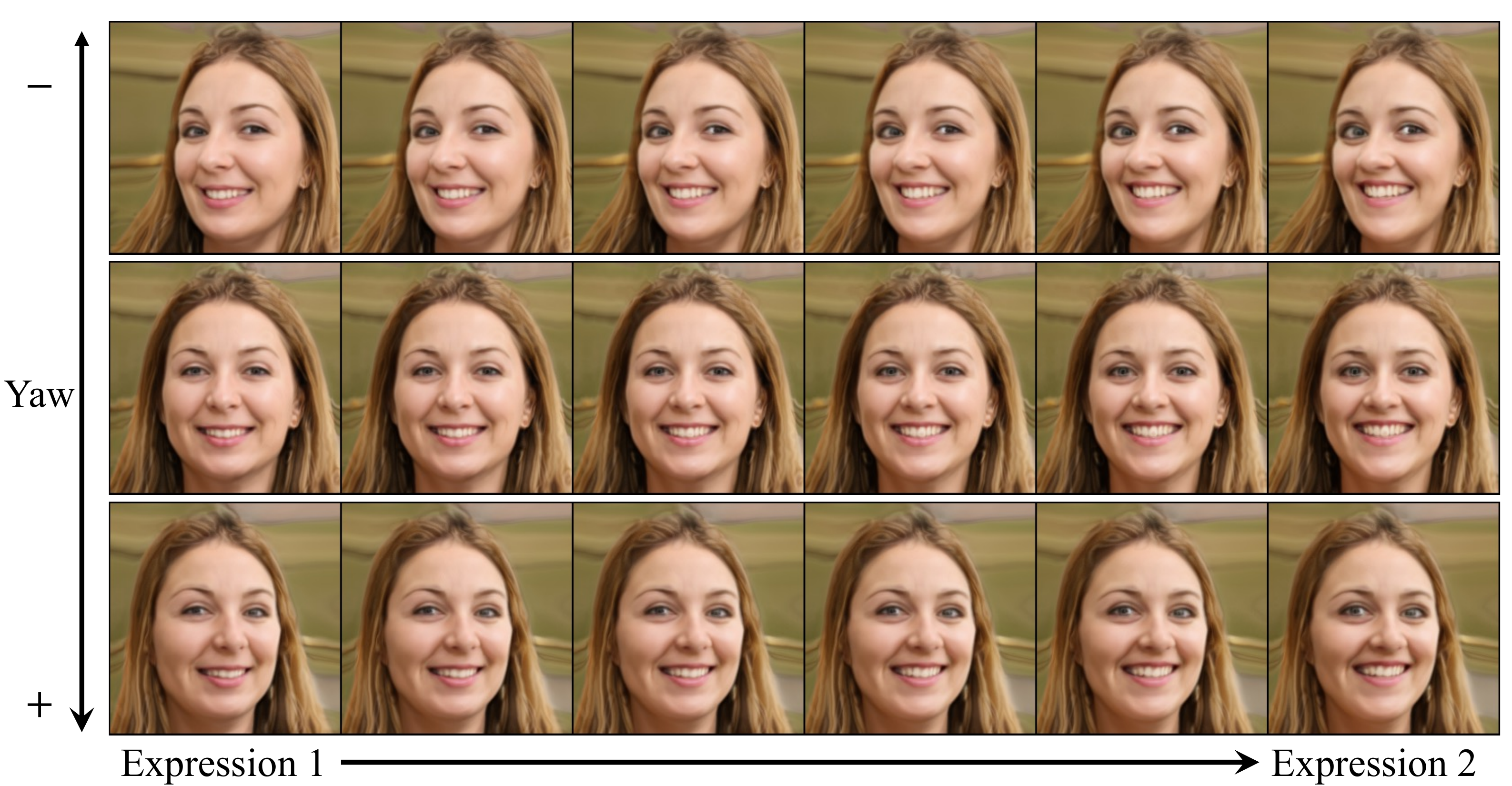}
\vspace{-.7cm}
\caption{\textbf{Multi-view motion interpolation in the latent space.} We linearly interpolate two motion representations during synthesis (\textit{column 1-6}), viewed from multiple directions (\textit{row 1-3}).}
\vspace{-.5cm}
\label{fig:interp}
\end{figure}

\noindent\textbf{Motion interpolation in the latent space.} 
To analyze the latent motion space, we render from multiple views a trajectory of synthetic images conditioned on linear interpolations between two motion representations, as shown in Fig.~\ref{fig:interp}. The smooth interpolation between two expressions over viewpoint changes demonstrates that the latent motion space captures semantics and CoRF enables to generate dynamic faces from multiple views with 3D-awareness.

\begin{figure}[t]
    \centering
\begin{minipage}{0.12\linewidth}
    \centering
    
    \cite{Wang_2020_CVPR}

     \vspace{1cm}
    \cite{tian2021a}
    
    \vspace{1cm}
    Ours
    \vspace{.1cm}
\end{minipage}
\begin{minipage}{0.86\linewidth}
\centering
\includegraphics[width=\linewidth]{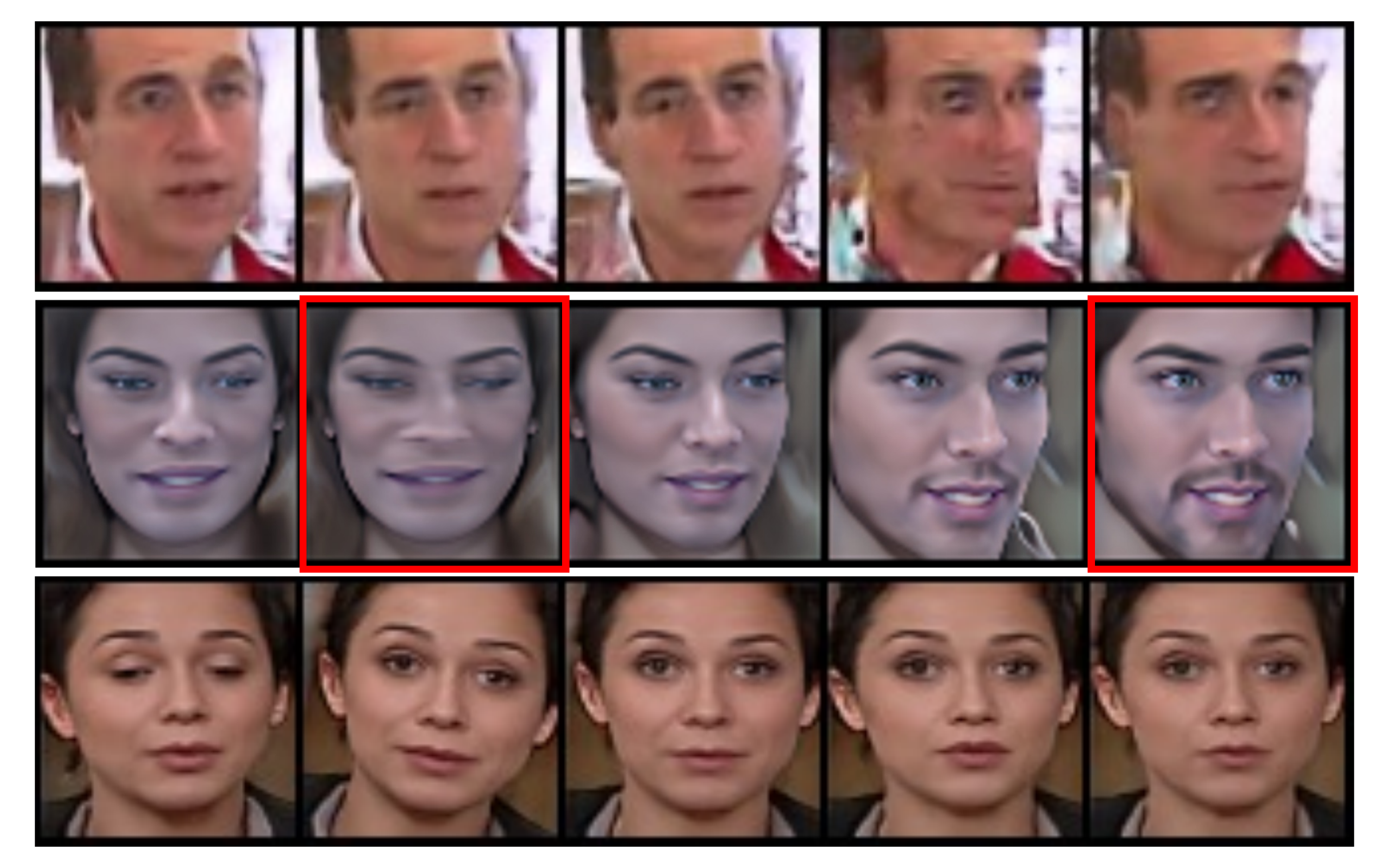}
\end{minipage}
\caption{\textbf{Visual comparison of video synthesis.} We compare our method with popular unconditional video synthesis approaches G$^3$AN ~\cite{Wang_2020_CVPR} and MoCoGAN-HD~\cite{tian2021a} on VoxCeleb2~\cite{vox2}. The  appearance of our method's results remains consistent over view and expression changes, while  baselines  struggle (see frames highlighted with red border). }

\label{fig:5}
\end{figure}

\noindent\textbf{Comparison to unconditional video synthesis methods.} We present a quantitative temporal coherence comparison in Tab.~\ref{tab:2} where we use 10,000 images/videos for numerical evaluation. Tab.~\ref{tab:2} shows that CoRF  significantly improves  video-level (FVD) quality compared to the baselines~\cite{ Wang_2020_CVPR, tian2021a}. 
In Fig.~\ref{fig:5}, we also show a comparison of unconditional video synthesis methods~\cite{ Wang_2020_CVPR, tian2021a} at a $64 \times 64$ resolution (size is chosen to be compatible with the baselines~\cite{ Wang_2020_CVPR, tian2021a}). We observe that CoRF produces sharper faces and persistent head geometry over time and space (\textit{row 3}).

\begin{figure}[t]
    \begin{minipage}[c]{\linewidth}
    \includegraphics[width=\linewidth]{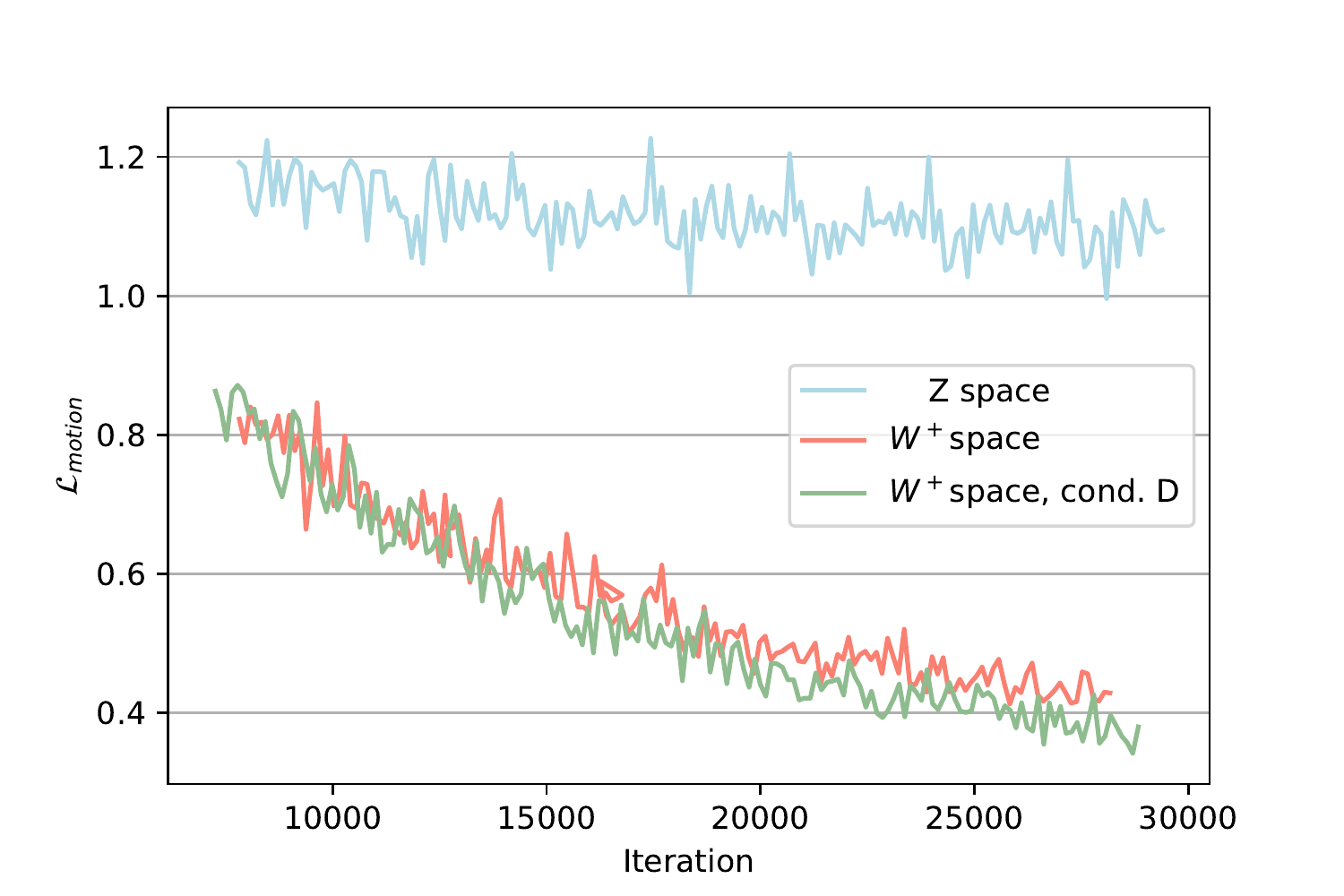}
    \end{minipage}
    
    \begin{minipage}[c]{\linewidth}
    \caption{\textbf{Analysis of motion conditioning strategies.} %
    \label{fig:4}
    }
    \end{minipage}
\end{figure}

\noindent\textbf{Ablation study.}
In Fig.~\ref{fig:4} we show the motion reconstruction loss curves during training with different motion conditioning strategies. The baseline method concatenates a motion representation $\bm m$ with a noise vector $\bm z$, %
as used in  prior work~\cite{gafni2020dynamic} (\textit{blue curve}). Compared to the baseline method, our motion embedding strategy (\textit{green and red curves}) accelerates  convergence of the motion reconstruction loss $\mathcal{L}_{\text{motion}}$. Moreover, we evaluate the effectiveness of conditioning  the discriminator on motion. We find that  motion conditioning of the discriminator (\textit{green curve}) is slightly better than omission of motion conditioning (\textit{red curve}). Thus, we use a conditional discriminator in all the experiments.

\begin{figure}[t]
    \centering
    \begin{minipage}{\linewidth}
    \centering
 
    \includegraphics[width=\linewidth]{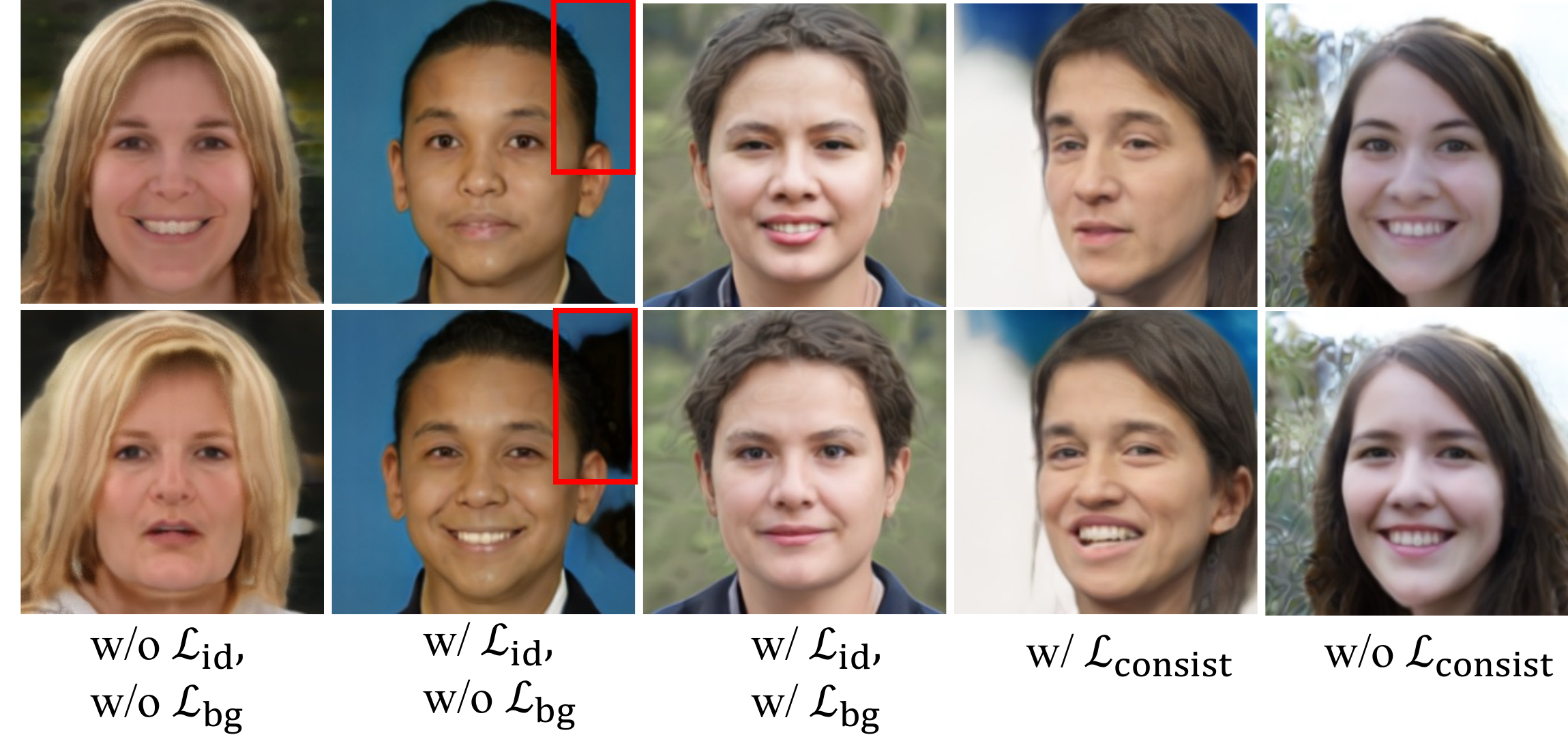}
    \end{minipage}
    \vspace{1cm} 
    
     \begin{minipage}{\linewidth}
        \centering
        \begin{tabular}{ccc}
        \toprule
            ~ &~$\mathcal{L}_\text{id}$~ & ~$\mathcal{L}_\text{bg}$~ \\
        \cmidrule{2-3}
          \makecell[c]{w/o $\mathcal{L}_\text{id}$, w/o $\mathcal{L}_\text{bg}$}  &0.07  &0.26 \\
          \makecell[c]{w/ $\mathcal{L}_\text{id}$, w/ $\mathcal{L}_\text{bg}$ }   &0.04 & 0.10 \\
          \bottomrule
        \end{tabular}
    \end{minipage}
    \caption{\textbf{Ablation study for losses.} We render paired synthetic images conditioned on the same identity representation but different motion representations (\textit{row 1-2}) using different ablation settings for identity, background, and consistency losses. In the table, we show averaged values of the losses $\mathcal{L}_\text{id}$ and $\mathcal{L}_\text{bg}$ (\textit{column 2-3}), without (or with) optimizing them.}
    \label{fig:abl}

\end{figure}

Fig.~\ref{fig:abl} provides a visual comparison for the ablation settings of the proposed consistency losses, illustrated below the figure (\textit{column 1-3}).
For each setting, we render two synthetic images applying an identical identity representation yet a different motion representation (\textit{top} and \textit{bottom}). We find that the identity loss $\mathcal{L}_\text{id}$, the background loss $\mathcal{L}_\text{bg}$, and the consistency loss $\mathcal{L}_\text{consist}$ help to preserve the face identity and a stable background over expression and pose changes. The table on the right shows the averaged loss values with or without optimizing the corresponding loss terms, which supports the visual observation.

\begin{table}[t]
    \centering
    \setlength{\tabcolsep}{3 pt}{
    \begin{tabular}{cccccccc}
    \toprule
    \centering
    ~ &\multicolumn{4}{c}{64 $\times$ 64} &~ &\multicolumn{2}{c}{256 $\times$ 256} \\
    \cmidrule(r){2-5} 
    \cmidrule(r){7-8} 
     ~ &~\cite{MoCoGAN} &~\cite{Wang_2020_CVPR}  &~\cite{TGAN2020}  
     &Ours  &~ &~\cite{tian2021a} & Ours\\
      \midrule
        FaceF~\cite{facef} &70.71& 65.58  &51.66 &\textbf{34.33}  &~ 
        & 53.26 & \textbf{35.67} \\
        Vox2~\cite{vox2} &40.06 & 36.18 &31.79 & \textbf{20.12}  &~ 
        &35.81 & \textbf{21.91} \\
     \bottomrule
    \end{tabular}
    }
    \caption{\textbf{Quantitative evaluation for video synthesis.} We evaluate our method using FVD~\cite{fvd} scores ($\downarrow$) and compare to MoCoGAN~\cite{MoCoGAN}, G$^3$AN~\cite{Wang_2020_CVPR},  TGANv2~\cite{TGAN2020} at the $64 \times 64$ resolution, 
    and with  MOCOGAN-HD~\cite{tian2021a} at the $256 \times 256$ resolution.}
    \label{tab:2}
\end{table}

\vspace{-.1cm}
\section{Conclusion}
\label{sec:05-con}
We propose controllable radiance fields (CoRFs) to address the task of generalizable, 3D-aware, and motion-controllable synthesis of face dynamics. 
CoRFs allow not only explicit 3D-aware motion control and arbitrary viewpoints but also diverse identity generation in photo-realistic quality. Extensive experiments highlight that CoRFs improve performance w.r.t.\ 3D consistency and identity preservation over pose and motion changes, compared to state-of-the-art methods on benchmark video datasets.  We hope this work  inspires future work on the challenging task of 3D-aware dynamic face synthesis.

\noindent\textbf{Acknowledgements.} This work is supported in part by NSF under Grants 1718221, 2008387, 2045586, 2106825, 1934986, MRI \#1725729, and NIFA award 2020-67021-32799. S.K. was supported by Google.
\newpage
{\small
\bibliographystyle{ieee_fullname}
\bibliography{final_main}
}

\clearpage
\appendix
\section{Additional experiments}
\label{app:add}

\subsection{Comparison of video generations}
\begin{figure}[h]
    \centering
     \begin{minipage}{0.07\linewidth}
     \centering
     \cite{MoCoGAN}
     \vspace{.5cm}
     
     \cite{TGAN2020}
     
     \vspace{1cm}
     Ours
      \end{minipage}
    \begin{minipage}{0.92\linewidth}
     \centering
    \includegraphics[width=\linewidth]{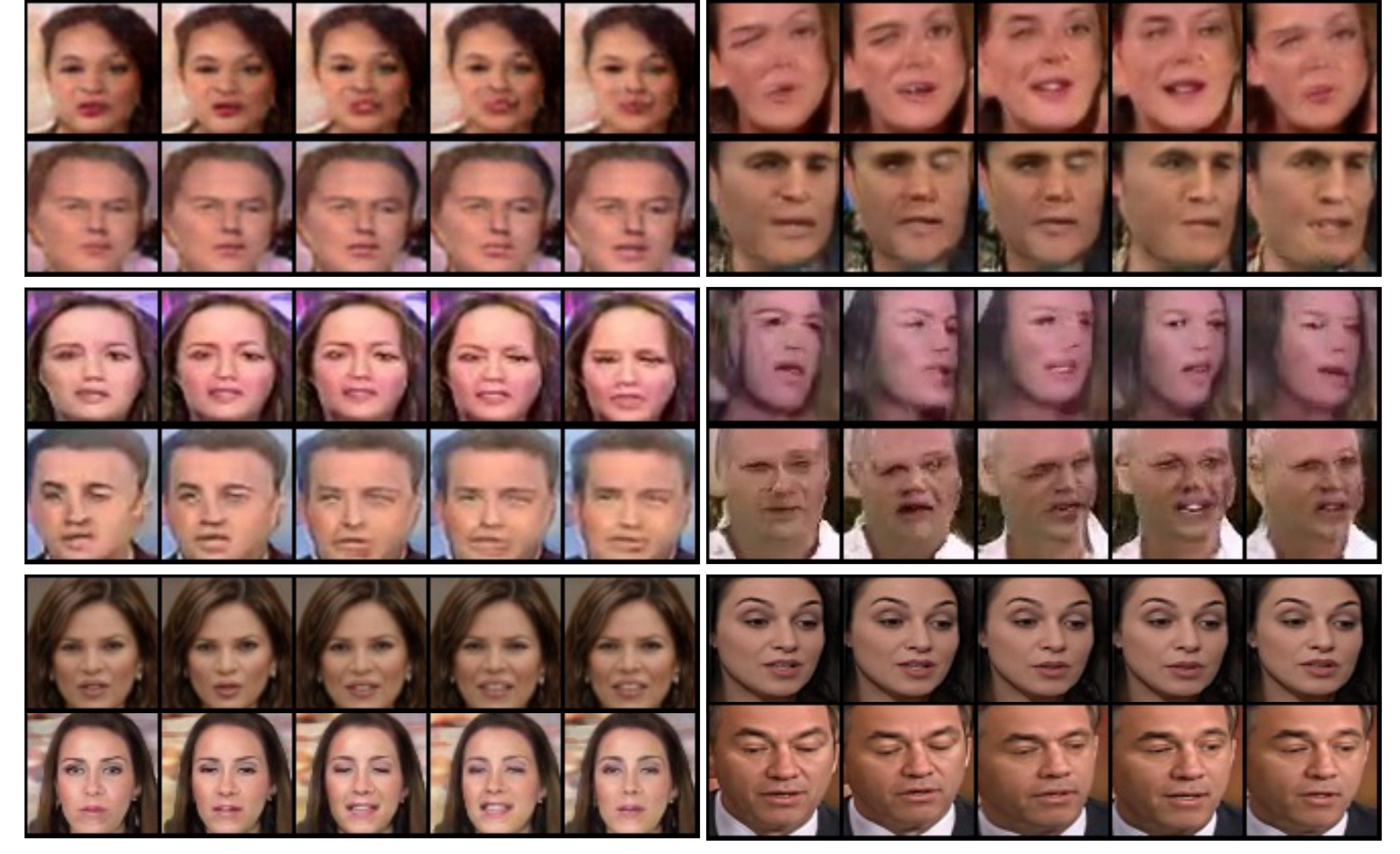}
    \end{minipage}
    
    \caption{\textbf{Video generation results of baseline methods.} We show videos generated by 2 baseline methods MoCoGAN~\cite{MoCoGAN} (\textit{row 1-2}) and TGANv2~\cite{TGAN2020} (\textit{row 3-4}), and ours (\textit{row 5-6}). }
    \label{fig:video_gen}
\end{figure}

We show synthetic videos from 2 baseline methods MoCoGAN~\cite{MoCoGAN} and TGANv2~\cite{TGAN2020}, and our CoRF at $64 \times 64$ resolution in Figure~\ref{fig:video_gen}.
\begin{figure}[h]
    \centering

    \begin{minipage}{\linewidth}
     \centering
    \includegraphics[width=\linewidth]{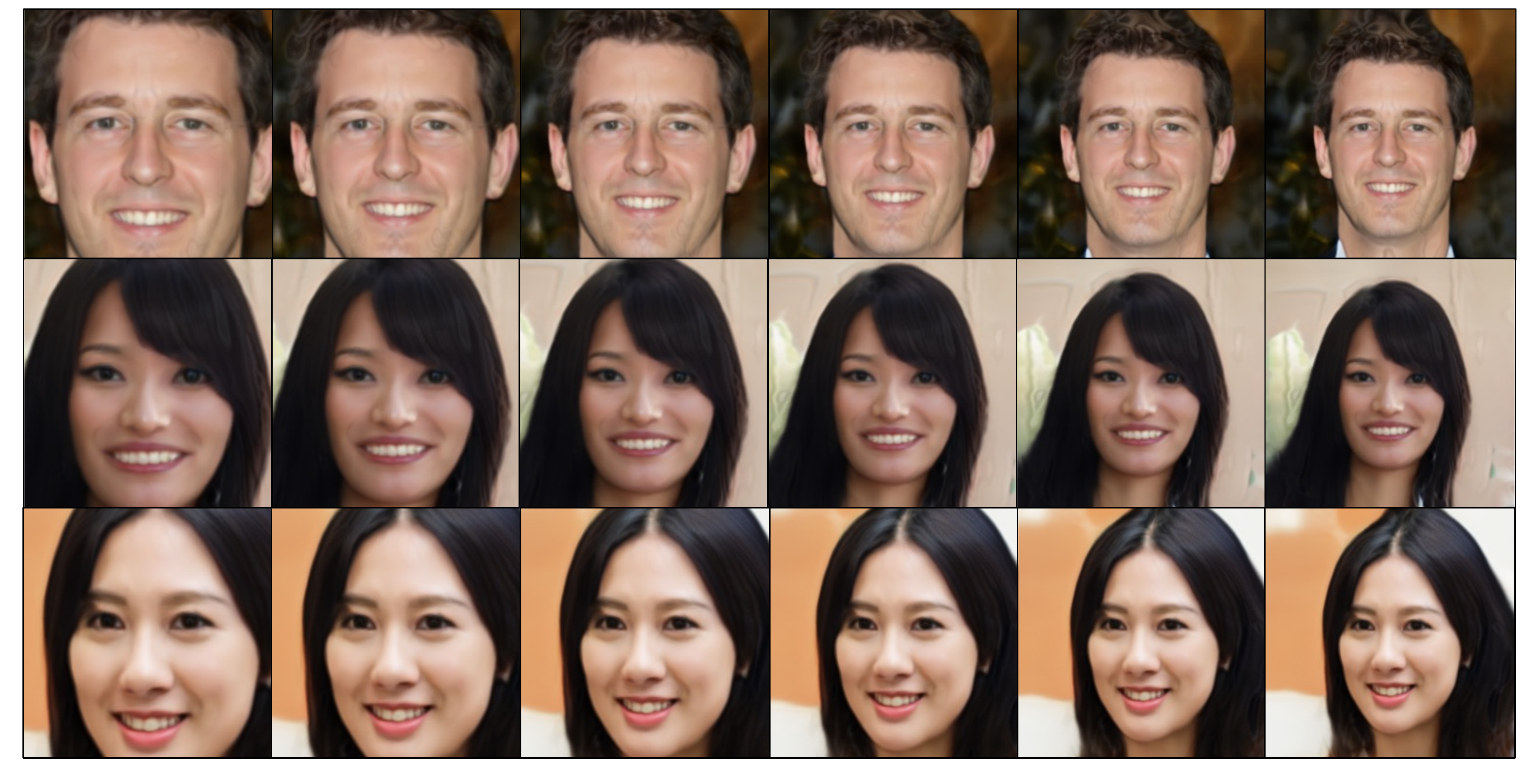}
    \end{minipage}
    
    \caption{\textbf{Rendered results under camera zoom-in to zoom-out.}}
    \label{fig:fov_sup}
\end{figure}

In Tab.~\ref{tab:std} we show the standard deviations of the evaluation metrics using MoCoGAN~\cite{MoCoGAN}, G$^3$AN~\cite{Wang_2020_CVPR}, TGANv2~\cite{TGAN2020} (corresponding mean values in Table 1 of the main manuscript), FOMM~\cite{FOMM} (corresponding mean values in Table 2 of the main manuscript), and our CoRF.

\begin{table*}[b]
    \centering
    \caption{\textbf{Standard deviations.} We show the standard deviations of KID, ID, EXP $^\dag$, EXP $\ddag$, and AKD scores using MoCoGAN~\cite{MoCoGAN}, G$^3$AN~\cite{Wang_2020_CVPR}, TGANv2~\cite{TGAN2020}, FOMM~\cite{FOMM}. The standard deviations are computed on the two datasets~\cite{facef, vox2} at $64 \times 64$ resolution (\textit{row 1-4}) and $256 \times 256$ resolution (\textit{row 5-6}).}
    \setlength{\tabcolsep}{2 pt}{
    \begin{tabular}{ccccccc|ccccc}
    \toprule
    \multirow{2}{*}{No.}
    &\multirow{2}{*}{Method}
    &\multicolumn{5}{c|}{FaceForensics++~\cite{facef}}  &\multicolumn{5}{c}{Vox2~\cite{vox2}}\\
    \cmidrule(r){3-12} 
     ~&~ & KID & ID & EXP $\dag$ & EXP $\ddag$ & AKD  & KID & ID & EXP $\dag$ & EXP $\ddag$ & AKD \\ 
      \midrule
    1 &\cite{MoCoGAN} (64) & 1$e$-3 &- &- &- &-
    & 1$e$-3 &- &- &- &- \\
    2 &\cite{Wang_2020_CVPR} (64) & 9$e$-4 &- &- &- &-
    & 1$e$-3 &- &- &- &-\\
    3 &\cite{TGAN2020} (64)  & 1$e$-3 &- &- &- &-
    & 9$e$-4 &- &- &- &- \\
    4 & Ours (64)    & 5$e$-4 &- &- &- &-
    & 5$e$-4 &- &- &- &-\\
    \midrule
    5 &\cite{FOMM} (256) &- & 4$e$-2 & 2$e$-1 &2$e$-1 & 2.2 &- &5$e$-2 & 2$e$-1 & 2$e$-1 & 2.5\\
    6 & Ours (256)  &- &2$e$-2 &1$e$-1 &1$e$-2 &2.5
    & - & 4$e$-2 & 1$e$-1 &1$e$-1 &2.9\\
    
     \bottomrule
    \end{tabular}}
    \label{tab:std}
\end{table*}
\begin{figure*}[b]
    \centering
     \begin{minipage}{0.15\linewidth}
     \centering
     \vspace{.5cm}
     Expression 1
     
     \vspace{1.5cm}
     COLMAP
     
     \vspace{1.5cm}
     Expression 2
     
     \vspace{1.5cm}
     COLMAP
      \end{minipage}
    \begin{minipage}{0.7\linewidth}
     \centering
    \includegraphics[width=\linewidth]{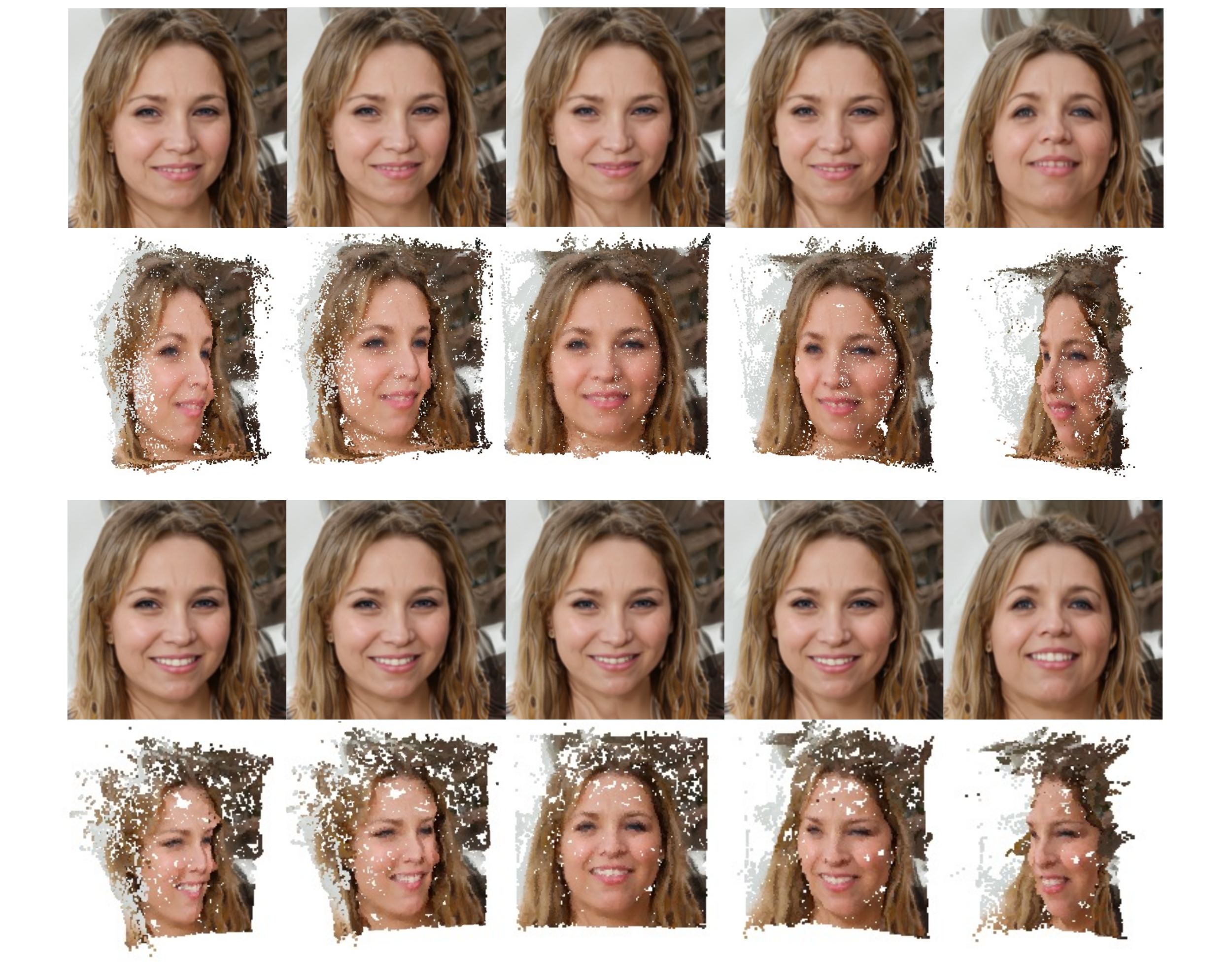}
    \end{minipage}
    
    \caption{\textbf{COLMAP reconstructions for models trained on FFHQ.} We show COLMAP reconstruction results (\textit{row 2 and 4}) for an identity with 2 different expressions (\textit{row 1 and 3}). }
    \label{fig:geo}
\end{figure*}
\subsection{3D reconstruction}
In Fig.~\ref{fig:geo} we show COLMAP~\cite{colmap} reconstruction results for an identity with 2 different expressions. Following~\cite{stylenerf}, we sample 36 camera poses to generate images and obtain the reconstructed point clouds from COLMAP with default parameters and without known camera poses.
\subsection{Camera zoom}
In Fig.~\ref{fig:fov_sup} we show that CoRF enables to generate reasonable results with camera zoom effects.

\section{Ethics discussion}
\label{app:ethic}

From an application perspective, we develop 3D-aware and motion-controllable image synthesis. We hope our method provides inspiration for generalization of NeRF-GAN models.
Importantly, we want to highlight the dangers of automated image manipulation. Similar to deepfake tasks whose aim is to produce fabricated images and videos that appear to be real. Improper use of image manipulation approaches might raise  issues with regard to information security, property, etc., and can be used for information fabrication or information surgery. Edited image detection techniques~\cite{wang2019cnngenerated} have been proposed recently to avoid the aforementioned issues, which promotes growth in both domains.

\end{document}